\documentclass[runningheads]{llncs}
\usepackage{graphicx}
\usepackage{amsmath,amssymb} 
\usepackage{color}
\usepackage[width=122mm,left=12mm,paperwidth=146mm,height=193mm,top=12mm,paperheight=217mm]{geometry}

\usepackage{epsfig}
\usepackage[caption=false,font=footnotesize]{subfig}

\usepackage{colortbl}
\definecolor{red}{rgb}{1,0,0}

\newcommand{\etal}{\textit{et al.}~}

\begin{document}

\pagestyle{headings}
\mainmatter

\title{Knowledge Transfer for Scene-specific Motion Prediction}

\titlerunning{Knowledge Transfer for Scene-specific Motion Prediction}
\authorrunning{L.~Ballan et al.}

\author{Lamberto Ballan\inst{1} \and Francesco Castaldo\inst{2} \and Alexandre Alahi\inst{1} \and Francesco Palmieri\inst{2} \and Silvio Savarese\inst{1}}
\institute{Computer Science Department, Stanford University
\email{\{lballan,alahi,ssilvio\}@stanford.edu}
\and Dept of Industrial and Information Engineering, Seconda Universit\`{a} di Napoli
\email{\{francesco.castaldo,francesco.palmieri\}@unina2.it}}

\maketitle


\begin{abstract}
When given a single frame of the video, humans can not only interpret the content of the scene, but also they are able to forecast the near future. This ability is mostly driven by their rich prior knowledge about the visual world, both in terms of (\emph{i}) the dynamics of moving agents, as well as (\emph{ii}) the semantic of the scene.
In this work we exploit the interplay between these two key elements to predict scene-specific motion patterns. First, we extract patch descriptors encoding the probability of moving to the adjacent patches, and the probability of being in that particular patch or changing behavior. Then, we introduce a Dynamic Bayesian Network which exploits this scene specific knowledge for trajectory prediction.
Experimental results demonstrate that our method is able to accurately predict trajectories and transfer predictions to a novel scene characterized by similar elements.
\end{abstract}

\section{Introduction}

Humans glance at an image and grasp what objects and regions are present in the scene, where they are, and how they interact with each other. But they can do even more. Humans are not only able to infer what is happening at the present instant, but also predict and visualize what can happen next. This ability to forecast the near future is mostly driven by the rich prior knowledge about the visual world.
Although many ingredients are involved in this process, we believe two are the main sources of prior knowledge: (\emph{i}) the \emph{static semantic of the scene} and (\emph{ii}) the \emph{dynamic of agents} moving in this scenario.
This is supported also by experiments showing that the human brain combines motion cues with static form cues, in order to imply motion in our natural environment \cite{Nature2003}.

\begin{figure}[!t]
\centering
\includegraphics[width=0.75\textwidth]{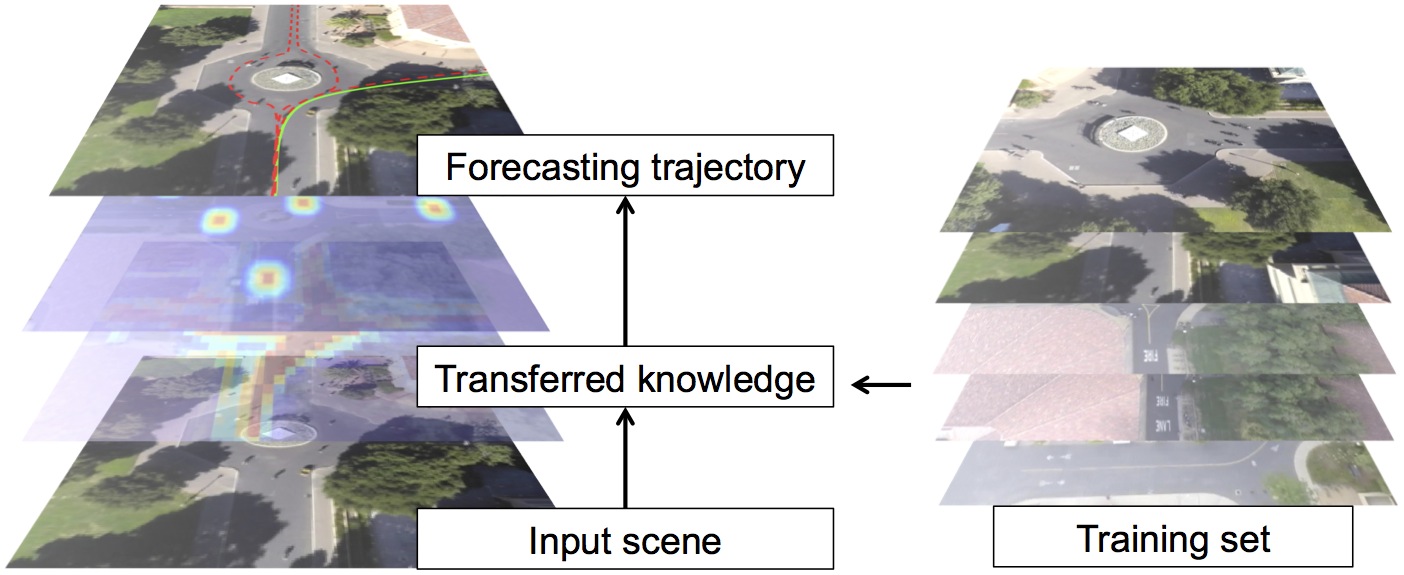}
\caption{Given the input scene shown in the bottom, we exploit the similarity between its semantic elements and those from a collection of training scenes, to enable activity forecasting (top image). This is achieved by transferring functional properties of a \emph{navigation map} that is learned from the training set. Such properties include local dynamic properties of the target, as well as typical route choices (middle).}
\label{fig:pullfig}
\end{figure}

Computer vision has a rich literature on analysing human trajectories and scenes, but most of the previous work addresses these problems separately.
Kitani \etal \cite{Kitani2012} have recently shown that by modeling the effect of the physical scene on the choice of human actions, it is possible to infer the future activity of people from visual data. Similarly, Walker \etal \cite{Walker2014} forecast not only the possible motion in the scene but also predict visual appearances in the future. Although these works show very interesting results, they considered only a few selected classes such as pedestrians and cars in a limited scenario.
This paper aims to take a step toward the goal of a generalized visual prediction paradigm, focusing on human-space (agent-scene) interaction.
Our main purpose is to take advantage of the interplay between the functional properties of the scene (instead of just semantic labels, such as grass, street, sidewalks) and the prior knowledge of moving agents, to forecast plausible paths from a new video scene.

We present a prediction model based on a Dynamic Bayesian Network formulation in which the target state is updated by using the statistics encoded in a \emph{navigation map} of the scene.
In training, the navigation map is formed as a collection of patch features, which encode the information about how previously observed agents of the same semantic class (\emph{e.g.}, pedestrian or cyclist) have moved from that particular patch to adjacent patches.
The class of the target is used to build different prediction models, since a pedestrian and a cyclist will probably navigate the same space in different manners.
We consider statistics that capture different properties: (\emph{i}) information about the direction and speed of the targets; (\emph{ii}) a score measuring how frequently that patch has been crossed; (\emph{iii}) identification of routing points, i.e. those patches in which the target is likely to turn or change behavior.
We call all these statistics \emph{functional properties} since they influence how an agent navigate the scene.
In testing, scene semantics are used to transfer functional properties from the training set to the test image, in order to build a navigation map on a novel scene. In turn, our model exploits the information encoded in the navigation map to drive the prediction.
Figure~\ref{fig:pullfig} illustrates our pipeline.

The contributions of this paper are three-fold:
(1) Our approach is able to infer the most likely future motion by taking into account how the targets interact with the surroundings.
In contrast to the previous work, our navigation map allows the model to predict rich navigation patterns by exploiting intermediate route points which lead to a more realistic path toward the final destination.
(2) The information encoded in our navigation map can be transferred to a novel scene that has never been seen before. 
(3) We show extensive results on path forecasting and knowledge transfer on multiple classes and a large set of scenes.

\section{Related Work}\label{related_work}

\textbf{Activity recognition and trajectory analysis.}
In activity recognition \cite{Wang2008,Zen2011,Amer2012} the main aim is to discover and label the actions of the agents observed in the scene. Depending on the level of granularity, we could be interested in atomic actions such as pedestrians walking in the scene, or in activities involving a group of people \cite{Lan10,alahi2009sparsity,Choi2012,ShuCVPR15,Solera2015}.
Trajectory-based activity analysis is a traditional way of modeling human activities \cite{Wang2008,Morris2008}, which leverages motion statistics to recognize different activities.
In this context, some recent work has shown that by directly modeling the impact of the environment, like trajectories of nearby pedestrians, can lead to better recognition models. To this end, social forces and contextual relationships have been effectively used to model human-human interactions in crowded environments \cite{social,Pellegrini2009,Yama2011,Alahi2014,Taixe2014,alahi2016lstm}.
Some other works show also that prior knowledge of goals yields better human-activity recognition and tracking \cite{Huang2008,Gong2011,xiang2015learning}.
From a complementary perspective, \cite{Wang2006,Turek2010} take advantage of trajectories to infer the semantic of the elements of the scene, such as road and sidewalk.

Our work falls in this broad area, but we focus on the interplay between the functional properties of the scene and the dynamic of agents observed in the scene. Moreover we are interested in using this joint observation of the scene to predict the future motion of our targets.
\smallskip

\textbf{Activity and trajectory forecasting.}
Some recent work \cite{Kitani2012,Fouhey2014,Xie2013} has put emphasis on \emph{predicting} unobserved future actions or, more generally, path forecasting \cite{Walker2014,Gavrila2013,Karasev2016}. Activity prediction greatly differs from recognition as in this case we do not have complete observations of the targets.
In their seminal work, Kitani \etal \cite{Kitani2012} have proposed an unified algorithm that learns the preferences of pedestrians navigating into the physical space. Their algorithm learns patterns such as a pedestrian prefers walking on the sidewalk avoiding obstacles (\emph{e.g.}, cars). However, a limitation of this model is that it can only model simple trajectories where the target goes from the initial point to a final destination, without considering any local intermediate goal.
More recently, Walker \etal \cite{Walker2014} presented a data-driven approach which exploits a large collection of videos to predict and visualize the most likely future frames. The main novelty of their approach is that it also yields a visual ``hallucination'' of probable events on top of the scene. 
However, their framework is strongly focused on predicting the visual appearance of the target and the results are mostly related to the single car-road scenario.
Similarly, \cite{Yuen2010,Vondrick2015} apply data-driven learning to leverage scene or video similarities to forecast the visual appearance of the near future frames. 
In our work we aim to make a step further by encoding agent-space interactions in our path prediction model. A key novelty of our approach is that the navigation map explicitly captures the functional interactions between the elements of the scene, and we show how this information is transferable to novel scenes.
\smallskip

\textbf{Exploiting objects functionalities.}
The elements constituting our environments have a strong impact in what kind of actions we usually do in these places. Object functionalities, also known as affordances, describe the possible interactions between an agent and an object (\emph{e.g.}, a chair is ``sittable'', an apple is ``eatable'', etc.).
Recently there has been growing interest in recognizing object and scene affordances in computer vision and human-robot interaction \cite{Gupta2009,Zhu2014,Saxena2015}.
A good example is given in \cite{Vu2014}, where action-scene correlations have been used for action prediction from static images.
Functional objects are regarded as ``dark matter'' in \cite{Xie2013}, emanating ``dark energy'' that can both attract (\emph{e.g.}, food-buses or vending machines) or repulse (\emph{e.g.}, buildings) humans during their activities.
This work is closely related to ours; however, they only study action-scene correlations, and the parameters learned on a specific scene are not transferable. In contrast, our model exploits the interconnection between the functional properties of the scene and the dynamic of agents moving in a particular context, to predict trajectories and transfer predictions to a novel scene.
\smallskip

\textbf{Knowledge transfer.}
Several works have shown that when a large training set is available, it is possible to use a simple approach based on data-driven visual similarity for retrieving and transferring knowledge to new queries in both image \cite{Hays2007,Liu2009,Tighe2010} and video \cite{Yuen2010,cviu2015,Gong2016} domains.
We build on this idea for transferring the knowledge encoded in our navigation map, relying on patch level similarities between our input scene and a training set of previously observed visual scenes.

\section{Path Prediction}\label{path_prediction}
\begin{figure*}[!t]
\centering   
\includegraphics[width=0.9\textwidth]{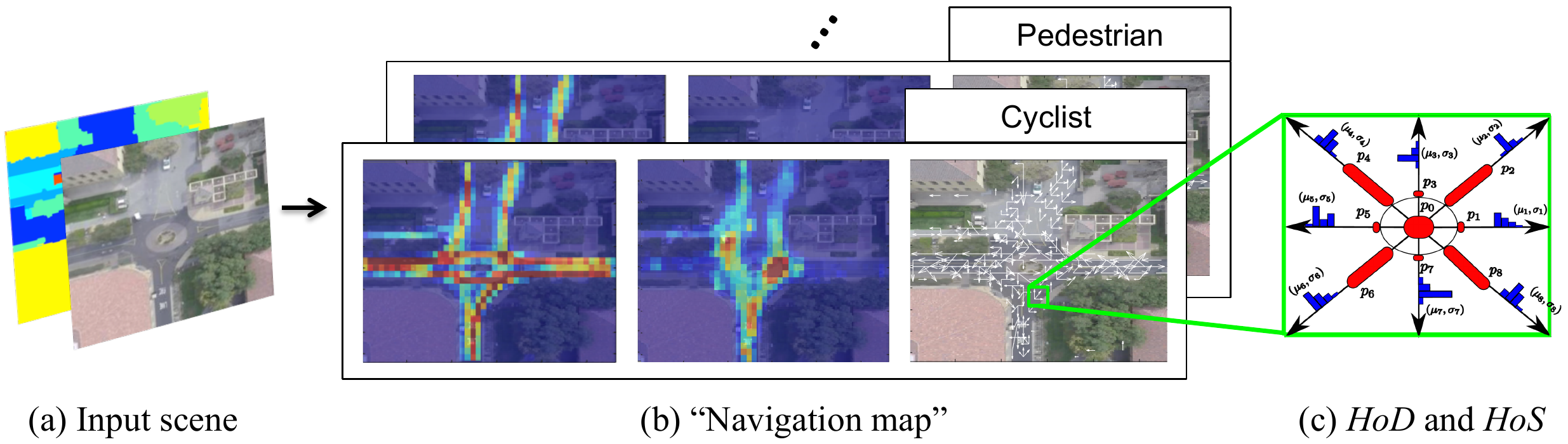}
\caption{We learn rich navigation patterns for each class of targets, and we leverage them to drive predictions. Given a scene (a), and a class label (\emph{e.g.}, cyclist, pedestrian), we collect a navigation map which captures the interactions between the target and the scene. (b) Shows, from left to right, popularity and routing scores, and Histograms of Directions and Speeds. (c) Visualizes HoD and HoS for a particular patch.}
\label{fig:navmap}
\end{figure*}

Figure~\ref{fig:pullfig} gives an overview of our approach.
Given a scene and a moving target, we aim at generating one (or more) likely trajectory of the target by using some prior information, namely the initial state of the target itself (position and velocity) and our knowledge of the scene.
The prediction is driven by what we call a \emph{navigation map}, which is described in detail in the following subsection and shown in Fig.~\ref{fig:navmap}.
The navigation map is a discrete representation of the space containing a rich set of information about navigation behaviors of targets that have been observed in training.
By integrating these statistics and the dynamics of targets, our prediction model allows the target to change its speed (if needed) and maneuver in the scene.
The class of the target is used to build different prediction models, since, for example, a pedestrian and a cyclist will navigate the same space in different manners. We achieve this goal in a probabilistic fashion, i.e. each predicted path is yielded by a underlying stochastic process, and has its own probability of being the ``real'' path.
It is important to note that our model relies on the functional properties of the scene. This is different from \cite{Kitani2012} in that they directly model the effect of the scene semantic on the human actions.
In contrast, in our model scene semantics are used only to transfer the navigation map features from a training set to a novel scene (see Section \ref{scene_transfer}).

\subsection{Navigation Map}\label{scene_partitioning}
Given an input scene, we overlay an uniform grid on it and build a map ${\cal M}$ which collects the navigation statistics. Given a class of targets, for each patch of the navigation map we consider four types of information, as described below.
\smallskip

\textbf{Popularity Score.}
The score $\rho \in[0,1]$ is related to the \emph{popularity} of each patch of the scene.
This score measures how many times a patch has been explored with respect to the others.
The popularity score can be used to pick the more likely paths, among all the solutions generated by our prediction model. The criteria is to favor paths crossing the highest number of popular patches. 

Figure \ref{fig:navmap}b (left) shows a heatmap of possible locations where a cyclist can be seen. It is quite easy to visualize the most common bike paths on the street.
\smallskip

\textbf{Routing Score.}
The score $\xi \in[0,1]$ is related to the probability of that patch of being a \emph{routing point}, that is a region in which the target is likely to turn or change its behavior, for instance near a turn or a wall (see the center image in Figure \ref{fig:navmap}b).
Those points can be viewed as the intermediate local goals the target crosses before heading to the final destination, and any path can be approximated to different linear sub-paths between such local goals.

The routing scores are calculated by evaluating the curvature values of the training paths over the map, i.e. by discovering in which patches the target significantly changes his behavior. The curvature $\cal K$ of a parametric curve in Cartesian coordinates  $c(t)=(x(t),y(t))$ is
\begin{equation}
{\cal K}=\frac{|\dot{x}\ddot{y}-\dot{y}\ddot{x}|}{(\dot{x}^2+\dot{y}^2)^{3/2}},
\end{equation}
where dots refer to first and second order derivatives with respect to $t$.
The routing values $\xi$ are then obtained by averaging the curvature values of each training trajectory, sampled at each patch.
\smallskip

\textbf{Histogram of Directions.}
The \emph{Histogram of Directions} (\emph{HoD}) represents the probability $p_i$, $i=0,\dots,N$, of heading  from there  into one of $N$ possible directions. In the following we use  $N=8$ to quantize the area in eight directions $\Theta_i$ (north, north-east, east, etc.), plus another fictitious direction $\Theta_{0}$ representing the possibility for the target to stop there. The distribution can easily account for not allowed directions (\emph{e.g.}, in cases in which there is a wall in that direction), by setting that probability to zero. 
\smallskip

\textbf{Histograms of Speeds.}
We compute $N$ \emph{Histograms of Speeds} (\emph{HoS}), each of them representing the expected velocity magnitude of targets leaving that patch (the velocity direction is already given in the \emph{HoS}). We represent the histograms by using $N$ Gamma distributions $\Gamma(\mu_i,\sigma_i$), because the data we are fitting are always positive and the support of the Gamma distribution is $(0,+\infty)$.
Figure \ref{fig:navmap}(c) shows an example of \emph{HoD} (in \emph{red}) and \emph{HoS} (in \emph{blue}).

\subsection{Prediction Model}
The target state variable is defined as ${\mathbf X}_k=({\mathbf P}_k,{\mathbf V}_k )^T$, at $k$th discrete time step, with ${\mathbf P}_k~=~(X_k,Y_k)^T$ (Cartesian position) and ${\mathbf V}_k~=~(\Omega_k,\Theta_k)^T$ (velocity magnitude and angle). The target interacts  with the navigation map $\cal M$, from which he extracts the navigation values (\emph{HoD}, \emph{HoS}, $\rho$ and $\xi$) for the patch he is occupying at each time. Starting from a given initial condition ${\mathbf X}_0$, our goal is to generate $T_p$ future states ${\mathbf X}_1,..{\mathbf X}_{T_p}$. A \emph{path}  $\Psi_{T_p}$ is defined as a collection $\{ {\mathbf X}_1,..,{\mathbf X}_{T_p}\}$ of target states.

The dynamic process describing the target motion is defined by the equations:
\begin{equation}
{\mathbf P}_{k+1}={\mathbf P}_k+\left(\begin{array}{c} \Omega_k  \cos  \Theta_k  \\ \Omega_k  \sin \Theta_k \end{array}\right)\Delta_k + {\mathbf w}_k,
\label{eq:ncv}
\end{equation}
\begin{equation}
{\mathbf V}_{k+1}=\Phi({\mathbf P}_k,{\mathbf V}_k;{\cal M}),
\label{eq:phi}
\end{equation}
where $\Delta_k$ is the sampling time (we assume $\Delta_k=1$) and ${\mathbf w}_k\sim {\cal N}({\mathbf 0},\sigma I_2)$ is a white-Gaussian process noise. Equation~\ref{eq:ncv} is a nearly-constant velocity model \cite{SurveyModel}, while Equation~\ref{eq:phi} represents the function which calculates the next speed vector ${\mathbf V}_{k+1}$, assuming we know the map $\cal M$ and the current state ${\mathbf X}_k$.
Although the nearly-constant velocity might seem a strong assumption (\emph{e.g.}, the agent may have a large acceleration at an intersection), we highlight that Eq.~\ref{eq:phi} and the learned expected values in $\cal M$ allows our model to generate non-linear behaviors.

Instead of trying to write a closed-form solution for $\Phi(\cdot)$, we resort to handling the process in probabilistic terms by means of a Dynamic Bayesian Network (DBN), where the target is modeled with a Gaussian distribution over its state.
The DBN is defined by the following conditional probability distributions:
\begin{equation}
p({\mathbf P}_{k+1}|{\mathbf P}_{k},{\mathbf V}_{k})={\cal N}(\left(\begin{array}{c} X_k +\Omega_k cos \Theta_k  \\ Y_k + \Omega_k sin \Theta_k  \end{array}\right),\sigma I_2),
\end{equation}
\begin{gather}
p({\mathbf V}_{k+1}|{\mathbf P}_{k},{\mathbf V}_{k};{\cal M})= 
\begin{cases}
p(\Omega_k |{\mathbf P}_{k},{\mathbf V}_{k};{\cal M}),   \\
p(\Theta_k|{\mathbf P}_{k},{\mathbf V}_{k};{\cal M}), \\
\end{cases}
\end{gather}
with 
\begin{equation}
p(\Omega_k |{\mathbf P}_{k},{\mathbf V}_{k};{\cal M})=\sum_{i=0}^N \Gamma(\mu_i,\sigma_i)p_{f_i},
\end{equation}
\begin{equation}
p(\Theta_k|{\mathbf P}_{k},{\mathbf V}_{k};{\cal M})=p_{f_i},
\end{equation}
where $\Gamma(\mu_i,\sigma_i)$ is the Gamma distribution which defines the $N$ speeds of the patch (we have chosen Gamma distribution  as it is not defined for negative values and has a single peak). The probability $p_{f_i}$ is written as

\begin{equation}
p_{f_i}=\frac{p_i e^{-\lambda d( \Theta_k, \Theta_i)}}{\sum_{j=0}^N p_j e^{-\lambda d( \Theta_k, \Theta_j)}},
\label{eq:pfi}
\end{equation}
where $d(\cdot)$ is a distance metric (here we use L2 norm) and $\lambda$ is a smoothing factor (the exponential is used to turn distances into probabilities). Equation~\ref{eq:pfi} means that we consider the similarity between the direction of the current speed vector $\Theta_k$ and all the other possible directions $\Theta_i$, weighted with the probability that the map assigns to that direction $p_i$.

\begin{figure}[!t]
\centering
\subfloat[Input scene]{\includegraphics[width=0.26\linewidth]{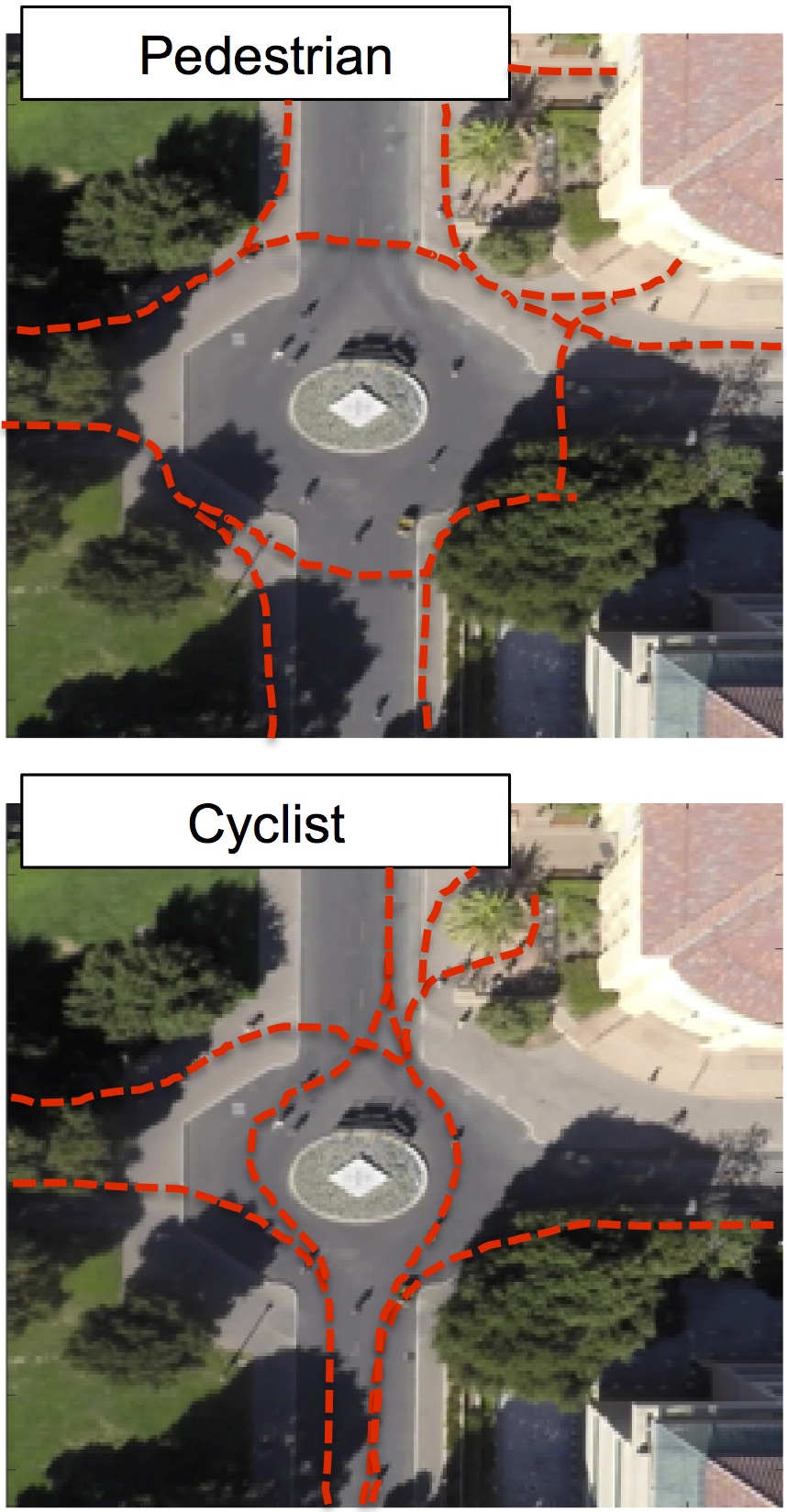}\hspace{1mm}}
\subfloat[Popularity map]{\includegraphics[width=0.26\linewidth]{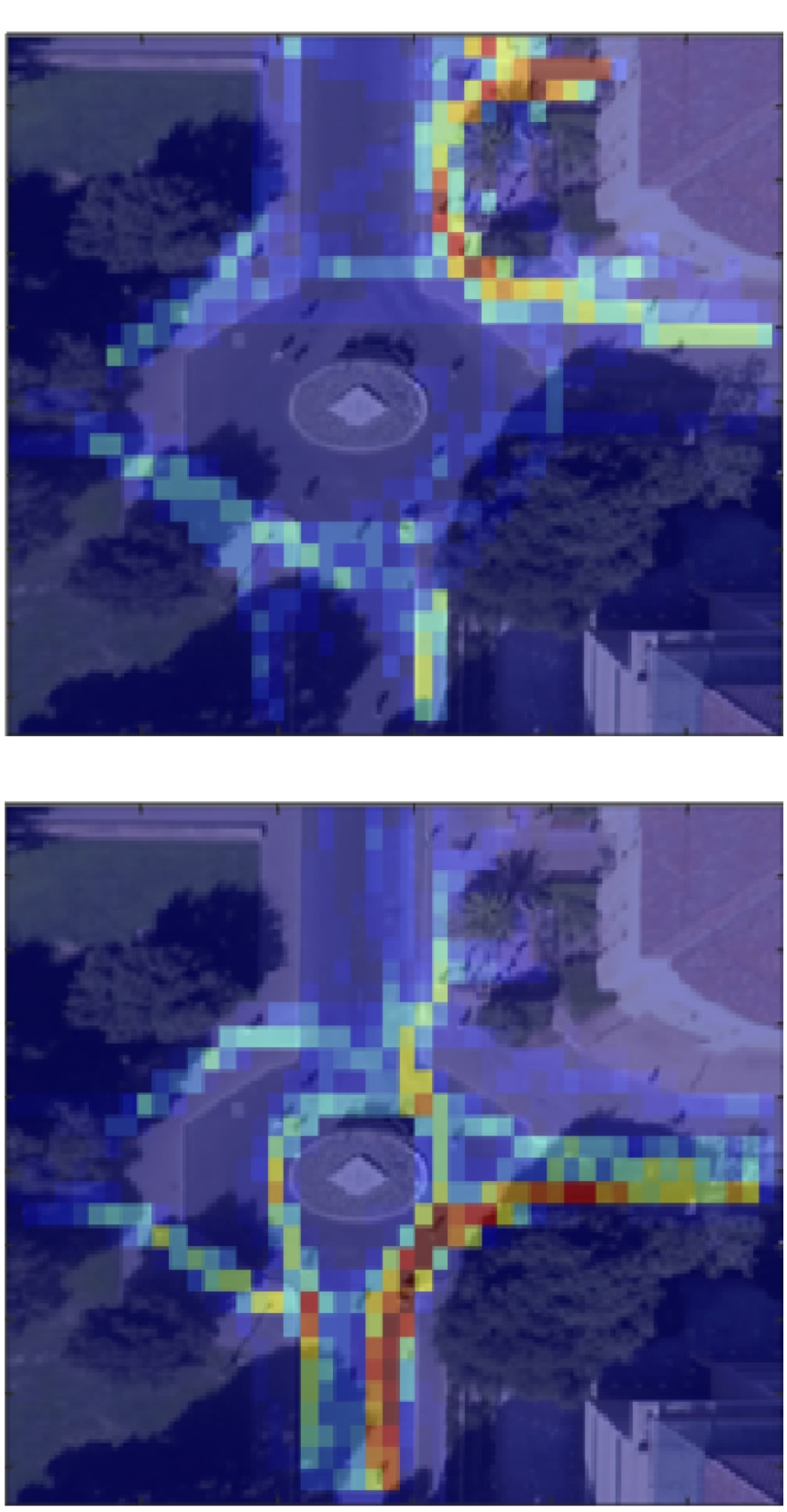}\hspace{1mm}}
\subfloat[Routing map]{\includegraphics[width=0.26\linewidth]{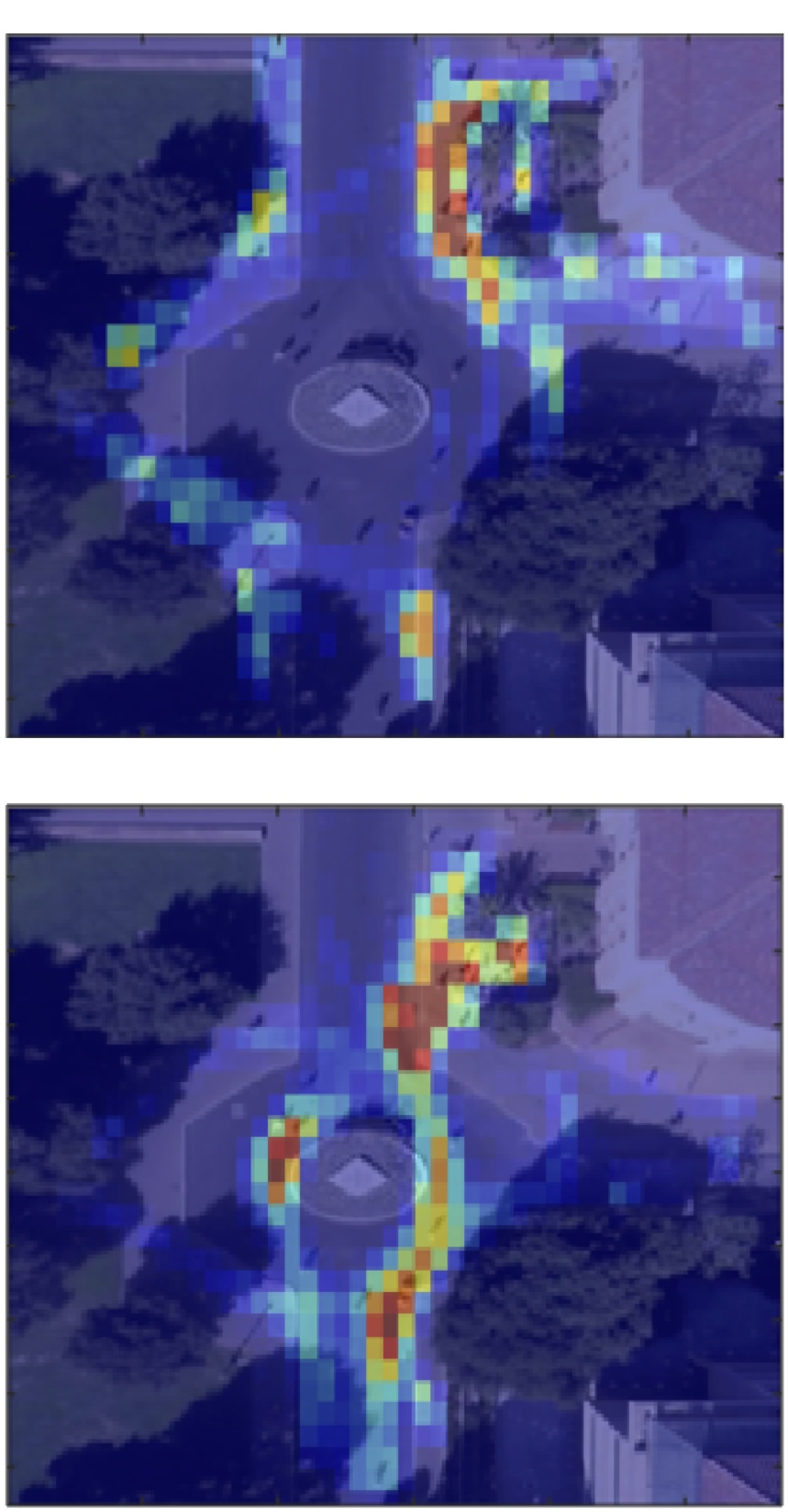}\hspace{1mm}}
\caption{Qualitative examples for two classes. Column (a) visualizes the most common paths for both classes on the same input scene. Columns (b) and (c) show the corresponding popularity and routing heatmaps. The cyclist map is particularly informative; it shows that the routing points are located in proximity of final destinations (\emph{e.g.}, the red area on top corresponds to bike racks) or areas where the agent may decide to turn.}
\label{fig:heatmaps}
\end{figure}

We need to modify the discrete values $p_{f_i}$ (which form a discrete distribution) described in Eq.~\ref{eq:pfi} in order to  incorporate the routing score $\xi \in[0,1]$.
Ideally, we want our distribution to be more ``randomic'' (i.e. more uniform) when the routing score $\xi$ is close to one, and more ``deterministic'' (i.e. always picking the most probable value) when $\xi$ tends to zero. These behaviors can be obtained with a Beta distribution $B(x;\alpha,\beta)\propto x^{\alpha-1}(1-x)^{\beta-1}$, with $\alpha=\beta$. It is easy to verify that the Beta distribution is uniform when $\alpha$ tends to zero and becomes a distribution peaking at the most probable value when $\alpha$ tends to infinity. For this reason  by writing the following transformation of random variable
\begin{equation}
\tilde{p_{f_i}}\propto p_{f_i}^\alpha  (1-p_{f_i})^\alpha,
\label{eq:gammabeta}
\end{equation}
with $\alpha=\frac{1-\xi}{\xi}$, we obtain the desired behavior and incorporate the routing score into the model. 

Given the probabilistic nature of the algorithm, by running it several times we get different paths. As explained in Section \ref{scene_partitioning}, we need a criteria to select the preferred path. We do so by leveraging the popularity score $\rho$ of each patch crossed by the target . The probability of the path $\Psi_{T_g}$ is calculated as
\begin{equation}
p(\Psi_{T_g})= \frac{1}{T_g} \sum_{i=1}^{T_g}  \rho_i,
\label{eq:popularity}
\end{equation}
where $\rho_i$ are the scores of the patches crossed by the target.
In our experimental section we report results using different strategies to generate the preferred path.

\section{Knowledge Transfer}\label{scene_transfer}

The activities of an agent, such as a pedestrian or a cyclist, are dependent on the semantic of the scene and the surroundings in which that particular agent appears.
The elements of the scene define a \emph{semantic context}, and they might determine similar behaviours in scenes characterized by a similar context.
So we design a simple retrieval-based approach which takes advantage of the similarity of the scene to transfer the functional properties (such as routing and popularity scores) that have been learned on the training set, to a new scene.
This is inspired by the success of previous nonparametric approaches for scene parsing \cite{Liu2009,Tighe2010,Yang2014}, but our goal is different since the local similarity of the scene is used to transfer the information required to build our navigation map.
This idea is also justified by the fact that we do not have a large dataset of scenes with both pixel-level labeling and trajectory annotations. Therefore, alternative approaches which requires intensive learning are not applicable in this scenario (\emph{e.g.}, end-to-end learning \cite{Farabet2013} or transferring mid-level representations \cite{Oquab14}).
\smallskip

\textbf{Scene parsing.}
The scene labeling is obtained using the scene parsing method presented in \cite{Yang2014}. For each image we extract several local and global features (SIFT + LLC encoding, GIST and color histograms). The algorithm first retrieves a set of image neighbors from the training set, then superpixel classification and MRF inference are used to refine the labeling~\footnote{We built on the code provided in~\cite{Yang2014}. The parameters to be set are the number of k-NN images ($9$ in our experiments) and of superpixels used to classify a segment (we used $5$). We used an intersection kernel and set the MRF pairwise term to $6$.}.
In order to evaluate how the quality of the scene labeling influences the performance of our model, we will present results using also the ground-truth image labeling. 
\smallskip

\begin{figure}[!t]
\centering
    \centering
    \subfloat[]{{\includegraphics[width=0.5\textwidth]{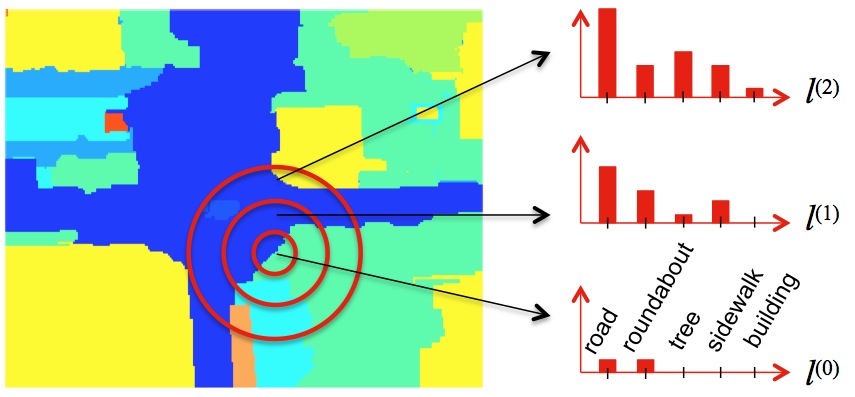}}}%
    \hspace{1mm}
    \subfloat[]{{\includegraphics[width=0.31\textwidth]{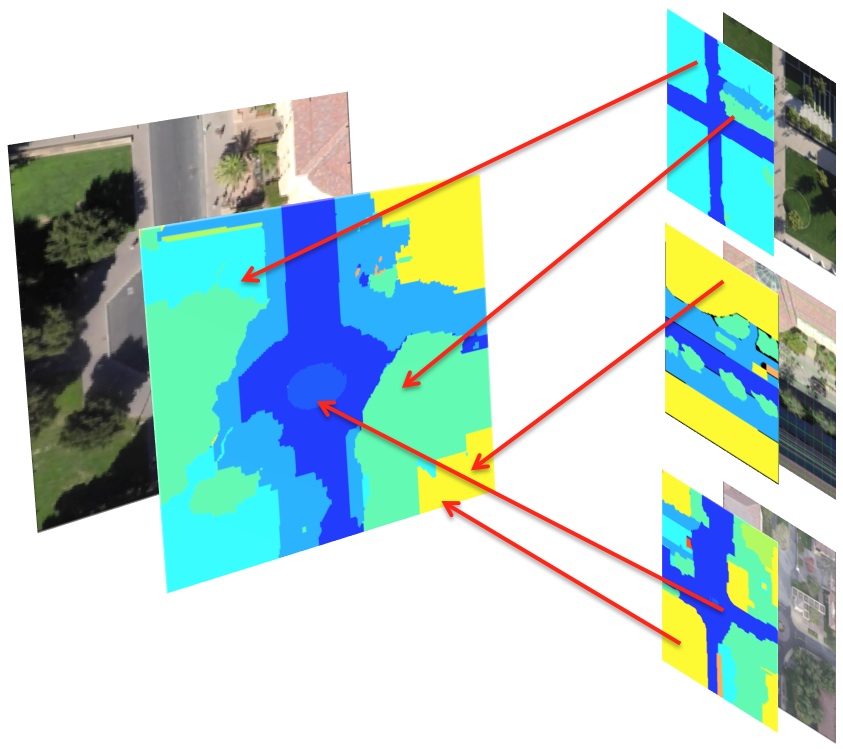}}}%
    \hspace{1mm}
    \subfloat[]{{\includegraphics[width=0.14\textwidth]{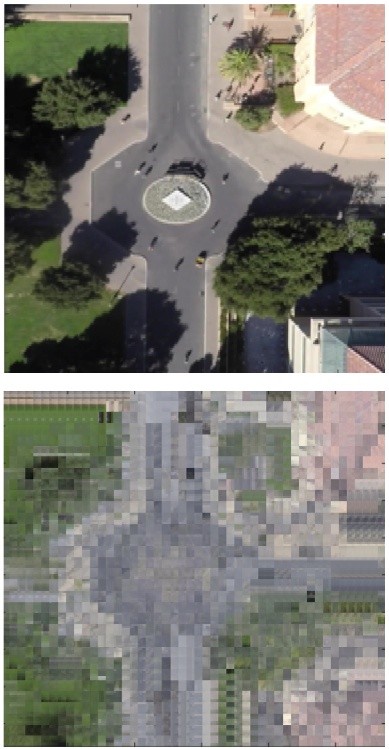}}}%
\caption{(a) Computing local context descriptors (three-levels). (b) Patch matching between an input scene and the training set; similarity is computed over the semantic patches. (c) Shows the input (top) and the \emph{hallucinated} scene (bottom) obtained by substituting each patch with its nearest-neighbors in the training set.}
\label{fig:cdm}
\end{figure}

\textbf{Semantic context descriptors.}
First, given a patch $i$, we define its \emph{global context}.
The global context descriptor is represented with a $C$-dimensional vector $\mathbf{g}_i$, where $C$ is the number of labels in the ground-truth. This is obtained by computing the Euclidean distance between the centroid of the patch and the closest point in the full image labeled as $c$, for each $c \in C$ (\emph{e.g.}, $c$ can be the class \emph{road}).
The role of the global context descriptor is to account for the relative distance between each patch, and all the other semantic elements of the scene. 

Then, we define the \emph{local context} of $i$ by looking at its contiguous local patches. We take inspiration from ``shape-context representations'' \cite{Belongie2002,Ballan2013}, and we define the local context $\mathbf{l}_i$ by encoding the spatial configuration of nearby patches at multiple levels (similarly to spatial pyramid).
We use a multilevel scheme, as illustrated in Figure~\ref{fig:cdm}(a). The space surrounding the current patch $i$ is partitioned into concentric shells, considering patches at distance 0, 1 and 2, in the grid.
For each patch, the local histograms are formed by counting the number of pixels labeled as $c$ in that shell. Thus we have multiple $C$-dimensional histograms (three in our example), computed at different levels. These histograms are then averaged, providing the final $\mathbf{l}_i$.
Note that this partitioning ignores absolute and relative orientations. We did some experiments considering also larger descriptors and a scheme with both sectors and bands, but this had a negative effect on performance.
The role of the local context descriptor is to account for the local ``arrangement'' of the scene, and it aims to capture the local similarities which might influence the agent's behaviours toward the intermediate goals.

The final patch descriptor $\mathbf{p}_i$ is a weighted concatenation of the global and local semantic context: $\mathbf{p}_i=w\mathbf{g}_i+(1-w)\mathbf{l}_i$, where $\mathbf{g}_i$ and $\mathbf{l}_i$ are L1 normalized.
The parameter $w$ is used to weight the contribution of the two components. Section~\ref{exp:semantic_transfer} shows how the performance is slightly influenced by this parameter.
\smallskip

\textbf{Descriptor matching.}
For each patch descriptor $\mathbf{p}_i$ in the query image, we rank all patches from training images using L2 distance and keep the set ${\cal N}_i$ of $K$ nearest-neighbors. Then, for each of the information collected for that patch (\emph{HoD}, \emph{HoS}, popularity and routing scores) we compute the average among the neighbors in ${\cal N}_i$ and we transfer this information to that particular patch $i$.
Intuitively, a good retrieval set contains semantic patches of similar scenes to the test image, in terms of both scene elements and spatial layouts.
Figure~\ref{fig:cdm}(b) illustrates the matching procedure, while Figure~\ref{fig:cdm}(c) shows a visualization of the ``hallucinated'' scene, obtained by substituting the original patches with their corresponding nearest-neighbors from the training set.

\begin{figure}[!t]
\centering
\includegraphics[width=0.49\textwidth]{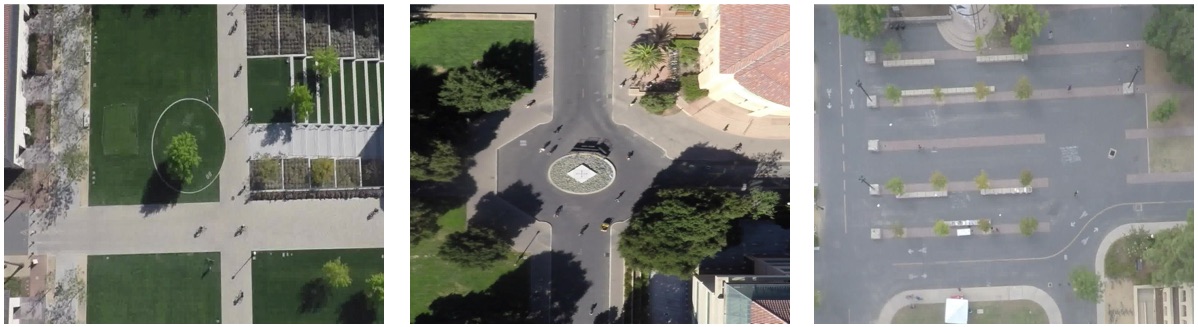}
\includegraphics[width=0.49\textwidth]{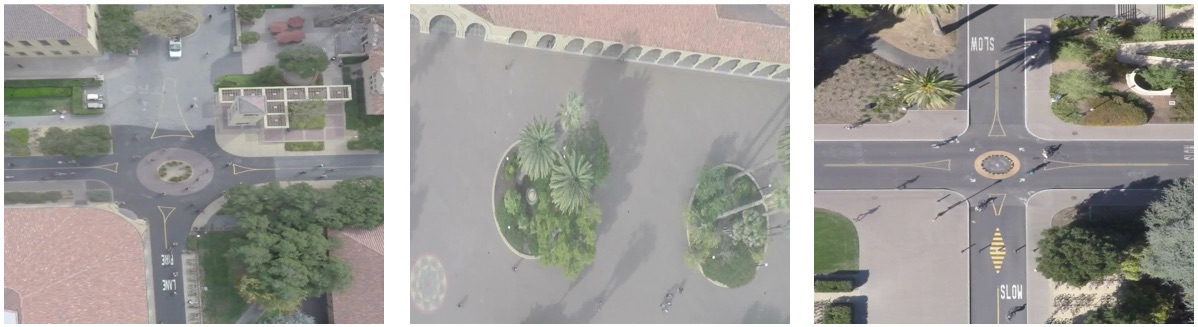}
\caption{Scenes from the Stanford-UAV dataset that has been recently presented in \cite{SUAV}.}
\label{fig:uav_dataset}
\end{figure}

\section{Experiments} \label{experiments}

\subsection{Dataset and Evaluation Protocol}
Previous works on visual prediction usually report results on a small number of videos with limited scene diversity (\emph{e.g.}, a subset of the VIRAT dataset \cite{virat}). Moreover, these datasets have been mostly used for video-surveillance applications and are limited to human activities or human-human interactions. In this work we use the UCLA dataset \cite{Amer2012}, along with a new challenging dataset that has been recently collected on Stanford campus from a UAV \cite{SUAV}.
\smallskip

\textbf{UCLA-courtyard dataset.}
The UCLA dataset \cite{Amer2012} consists of six annotated videos taken from two viewpoints of a courtyard at the UCLA campus.
Different human activities can be spotted in these videos (people moving back and forth, entering and exiting the scene, ordering food at the food bus, talking with each other, etc.). Although the dataset has been originally collected for recognizing group activities, the semantic of the scene is quite rich.
In order to allow the computation of our navigation map and the approach presented in \cite{Kitani2012}, we have manually labelled the scene with $8$ semantic classes: \emph{road}, \emph{sidewalk}, \emph{pedestrian}, \emph{car}, \emph{building}, \emph{grass}, \emph{food bus}, \emph{stairs}, \emph{bench}.
\smallskip

\textbf{Stanford-UAV dataset.}
This new dataset \cite{SUAV} provides many urban scenes (intersections, roundabouts, etc.) and several distinct classes of agents (pedestrians, cars, cyclists, skateboarders, baby carriages). The data have been collected from UAV. All videos are stabilized by registering each video frame onto a reference plane. In total we use $21$ videos from $6$ large physical areas (approx.~$900m^2$/scene), corresponding to $15$ different scenes.
In our experiments we consider only two target classes, namely \emph{pedestrian} and \emph{cyclist}, because for other classes the number of trajectories available in the dataset is too scarce.
We provide also the scene labeling of each scene for $10$ semantic classes: \emph{road}, \emph{roundabout}, \emph{sidewalk}, \emph{grass}, \emph{tree}, \emph{bench}, \emph{building}, \emph{bike rack}, \emph{parking lot}, \emph{background}.
\smallskip

\textbf{Evaluation metric.}
We use the Modified Hausdorff Distance (MHD) \cite{Dubuisson94amodified} to measure the pixel distance between ground-truth trajectories and estimated paths (as in previous work \cite{Kitani2012,Walker2014}).
In our evaluation, a path is assumed over when the target goes out of the scene, or when it reaches its final destination.

\subsection{Path Forecasting}
\label{exp:path_forecasting}
First, we test the performance of our framework on path forecasting on both UCLA-courtyard and Stanford-UAV dataset.
We compare with Kitani \etal \cite{Kitani2012}, which is based on inverse optimal control, and with a simple linear constant-velocity prediction baseline (similarly to \cite{Gavrila2013,Karasev2016}). The latter is the simplest version of our model when it does not leverage the information encoded in the navigation map.
Additionally, on the Stanford-UAV dataset we also run a social-force model (similarly to \cite{Yama2011}) and a linear prediction baseline with collision avoidance.
Each method will generate a collection of likely predicted paths. In both datasets, we use $70\%$ of the data for training and the rest for testing; results are reported with $5$-fold cross validation.
In the rest of this section we often refer to the linear prediction baseline as LP, and to Kitani \etal \cite{Kitani2012} as IOC.
\smallskip

\begin{figure}[!t]
\centering
{\includegraphics[width=0.2631\linewidth,height=0.22\linewidth]{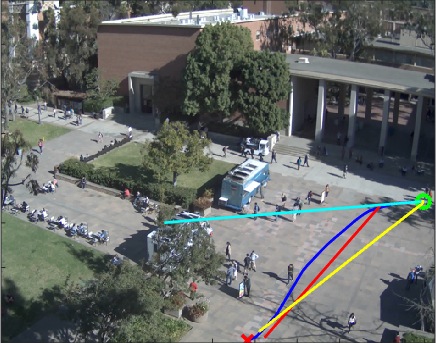}\hspace{.5mm}}
{\includegraphics[width=0.2631\linewidth,height=0.22\linewidth]{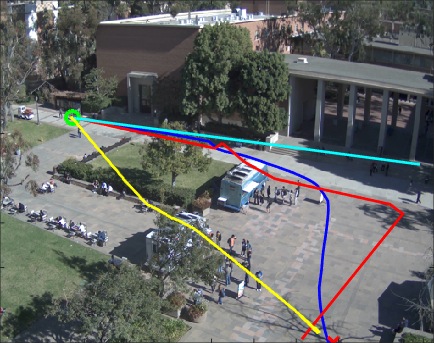}\hspace{.5mm}}
{\includegraphics[width=0.2631\linewidth,height=0.22\linewidth]{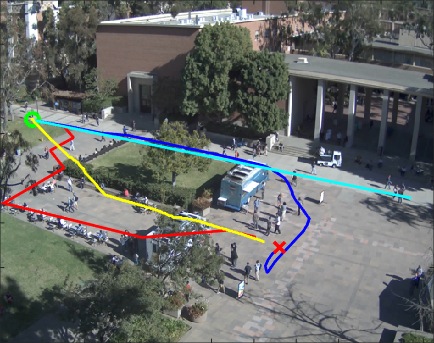}\hspace{.5mm}}
\caption{Qualitative results on the UCLA dataset. The \emph{blue} trajectory is ground-truth; \emph{cyan} is the linear-prediction (LP) baseline; \emph{yellow} is IOC \cite{Kitani2012}; \emph{red} is our model.}
\label{fig:ucla_exp}
\end{figure}

\textbf{Qualitative results.}
Figure \ref{fig:ucla_exp} shows some qualitative results. We observe that the trajectories predicted by our algorithm (in red) are usually very close to the ground truth paths (in blue).
With respect to IOC \cite{Kitani2012} (yellow paths), we can observe that our algorithm is better in capturing the human preference in the navigation of the scene.
This is particularly evident in the example shown in the middle, where the path predicted by our algorithm is significantly closer to the ground truth trajectory.
This example highlights the main properties of our model. While IOC \cite{Kitani2012} aims to directly optimize the path leading to the final destination, our model tends to describe complex patterns in which the target stops to intermediate goals before heading to its final destination.
\smallskip

\begin{table}[!ht]
\caption{(a) Path forecasting results on both datasets; we report the mean MHD error of the closest path for a given final destination. (b) Shows the results of our method on the Stanford-UAV dataset, obtained using different path generation strategies.}
\centering
\subfloat[Path forecasting]{
\begin{tabular}[b]{c||c|c}
\multicolumn{3}{c}{MHD error} \\
\cline{1-3}
\hline
& \emph{UCLA-courtyard} & \emph{Stanford-UAV}\\
\hline\hline
LP & $41.36 \scriptstyle{\pm 0.98}$ & $31.29 \scriptstyle{\pm 1.25}$\\
LP$_{CA}$ & - & $21.30 \scriptstyle{\pm 0.80}$\\
IOC~\cite{Kitani2012} & $14.47 \scriptstyle{\pm 0.77}$ & $14.02 \scriptstyle{\pm 1.13}$\\
SFM~\cite{Yama2011} & - & $12.10 \scriptstyle{\pm 0.60}$\\
\hline
Ours & $\mathbf{10.32} \scriptstyle{\pm 0.51}$ & $\mathbf{8.44} \scriptstyle{\pm 0.72}$\\
\hline
\end{tabular}
} \quad
\subfloat[Path generation (Ours)]{
\includegraphics[width=0.4\textwidth]{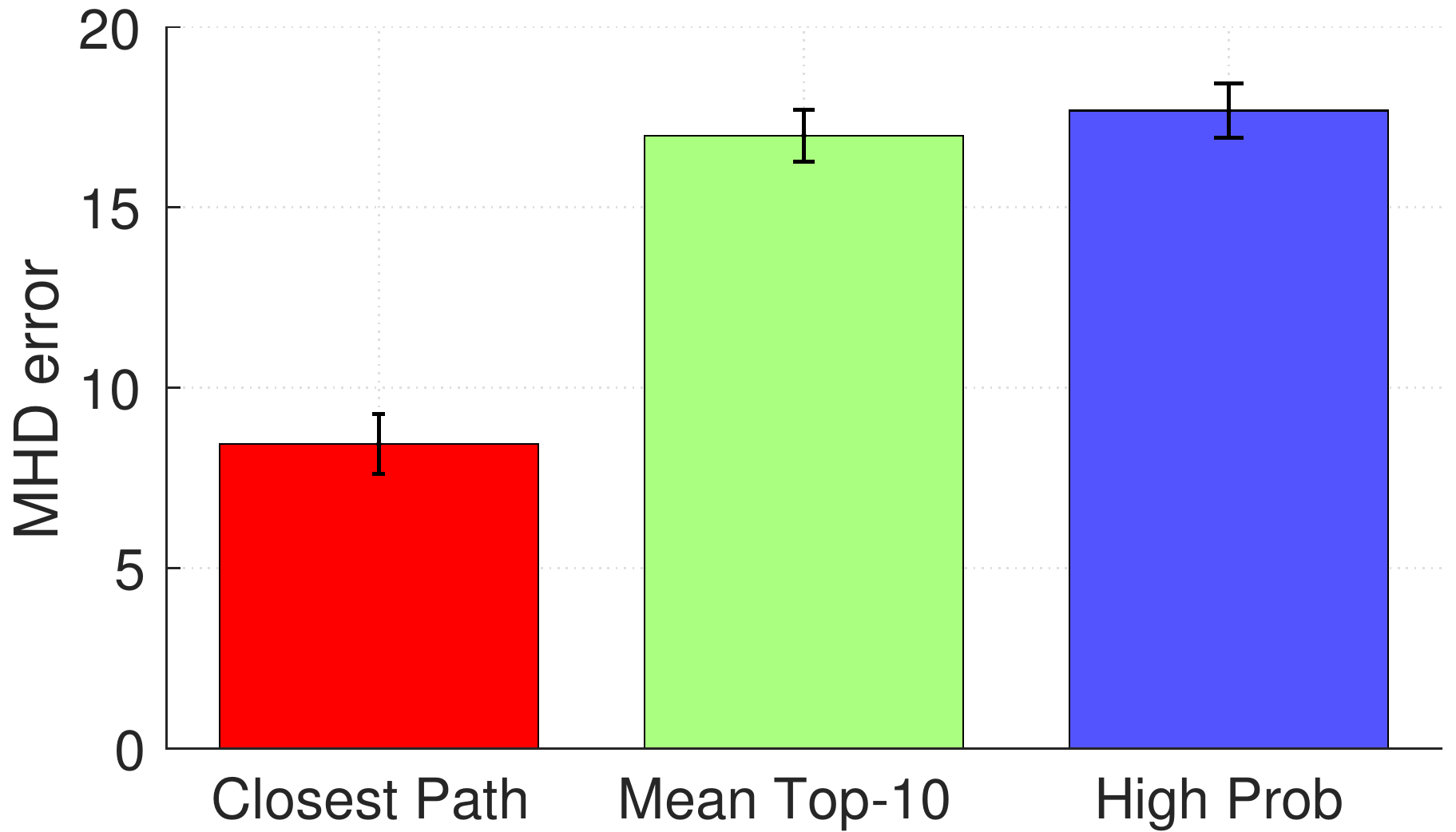}
}
\label{table1}
\end{table}

\textbf{Quantitative results.}
We report results on path forecasting in Table \ref{table1}. All the results in (a) are obtained providing both initial point and final destination, and refer to UCLA and Stanford-UAV dataset.
Our model significantly outperforms the linear prediction baseline (LP), and also IOC \cite{Kitani2012}.
Additionally, we report results on the Stanford-UAV dataset using a linear prediction baseline with collision avoidance (LP$_{CA}$), and the social force model \cite{Yama2011} (referred as to SFM) which models both human-space and human-human interactions.
These results confirm the effectiveness of our approach even when it is compared to other methods which take advantage of the interactions between agents.

In Table \ref{table1}(b) we show some results in which we investigate different path generation strategies. In other words, this is the strategy we use in our model to predict the final path among the most likely ones (see Eq.~\ref{eq:popularity}).
We obtained the best results when we privilege a path in which the final point is closest to the goal, but significant improvements can be obtained also if we peak the path with the highest popularity scores, or the mean of the top-10 most probable paths.

\subsection{Knowledge Transfer}
\label{exp:semantic_transfer}

\begin{figure}[!b]
\centering
\subfloat[Original image]{\includegraphics[width=0.23\linewidth]{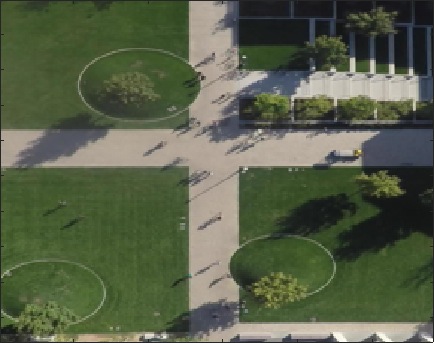}\label{fig:pt0}\hspace{1mm}}
\subfloat[K=5]{\includegraphics[width=0.23\linewidth]{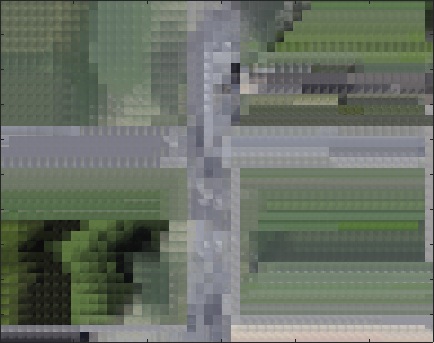}\label{fig:pt5}\hspace{1mm}}
\subfloat[K=10]{\includegraphics[width=0.23\linewidth]{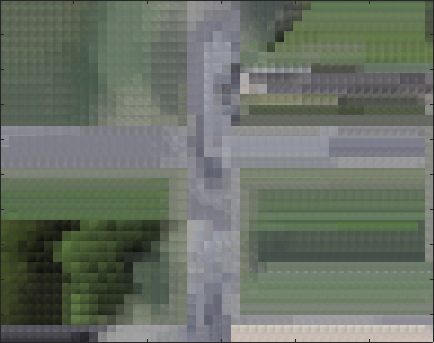}\label{fig:pt10}\hspace{1mm}}
\subfloat[K=50]{\includegraphics[width=0.23\linewidth]{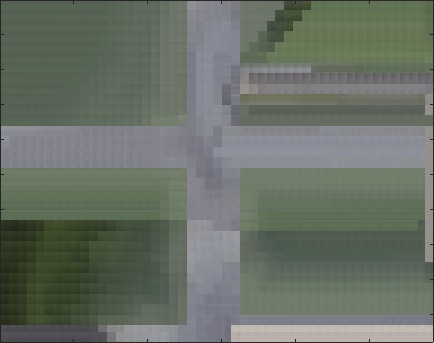}\label{fig:pt50}\hspace{1mm}}
\caption{(a) Input scene. (b,c,d) Show the ``hallucinated'' scene computed using our patch matching approach. The images are formed by average patches obtained with an increasing number of neighbors K. We varied the parameter $K$ in the interval $\lbrack1,200\rbrack$.}%
\label{fig:knn_patches}%
\end{figure}

Here we evaluate the ability of our model to generalize and make predictions on novel scenes.
This generalization property is very important since it is hard and expensive to collect large statistics of agents moving in different scenes.

Some preliminary experiments on knowledge transfer have been presented also in \cite{Kitani2012}, but they limited their study to a few different scenes, while we conduct an extensive analysis on the Stanford-UAV dataset.
By looking at the examples in Fig.~\ref{fig:uav_dataset}, we see that many elements in the scene, such as roundabouts or grass areas between road intersections, may often appear in similar configurations. Those regularities across the scenes are detected by our semantic context descriptors, and transferred by our retrieval and patch matching procedure.
\smallskip

\begin{figure}[!t]
\centering
\subfloat[Input scene]{
\begin{tabular}{c}
\includegraphics[width=0.2025\linewidth]{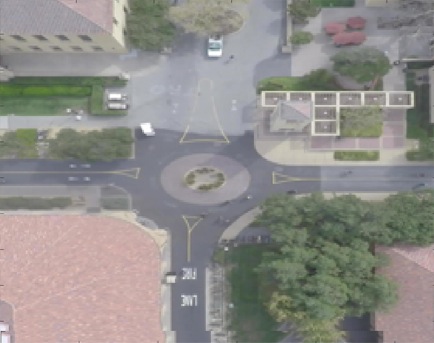}\\
\includegraphics[width=0.2025\linewidth]{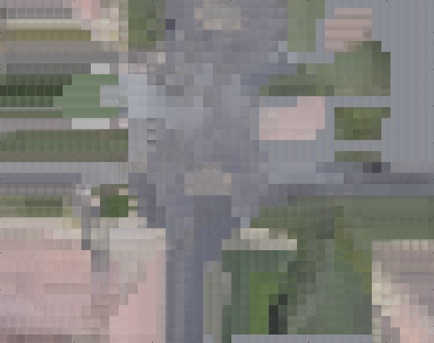}
\end{tabular}\hspace{-2mm}
}
\subfloat[Popularity map]{
\begin{tabular}{c}
\includegraphics[width=0.2025\linewidth]{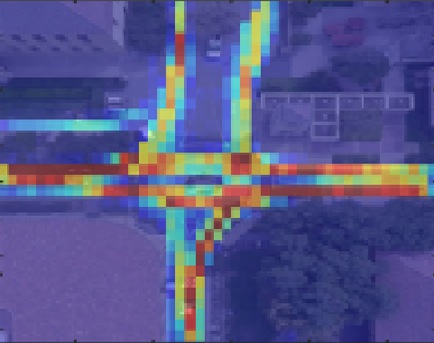}\\
\includegraphics[width=0.2025\linewidth]{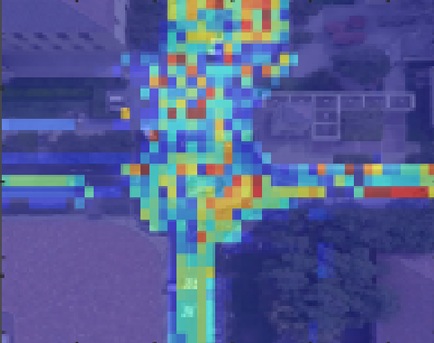}
\end{tabular}\hspace{-2mm}
}
\subfloat[Routing map]{
\begin{tabular}{c}
\includegraphics[width=0.2025\linewidth]{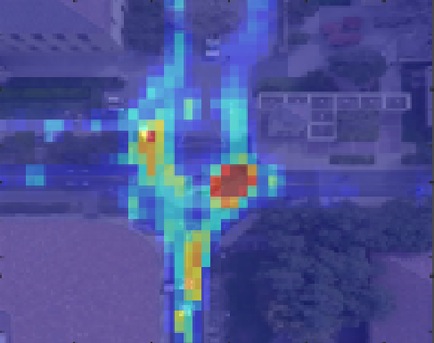}\\
\includegraphics[width=0.2025\linewidth]{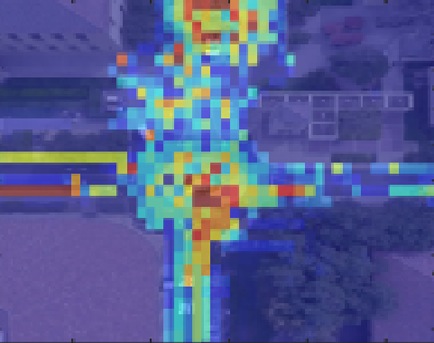}
\end{tabular}\hspace{-2mm}
}
\subfloat[Path prediction]{
\begin{tabular}{c}
\includegraphics[width=0.2025\linewidth]{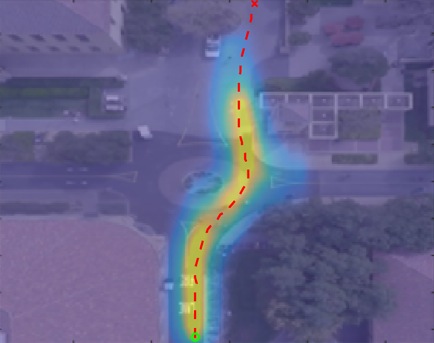}\\
\includegraphics[width=0.2025\linewidth]{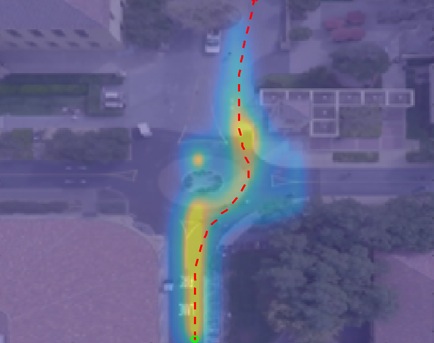}
\end{tabular}\hspace{-2mm}
}
\caption{The first row shows the results obtained in a standard path forecasting setting, while the second row shows results obtained after knowledge transfer. (a) Input scenes; (b,c) popularity and routing heatmaps; (c) demonstrates the predicted trajectory.}
\label{fig:transf_final}%
\end{figure}

\textbf{Qualitative results.}
Figure \ref{fig:knn_patches} shows a qualitative example of an ``hallucinated'' scene, obtained substituting each patch of the new scene with the most similar ones from the training set. Increasing the number of nearest-neighbors $K$, we can observe more coherent structures.
The actual knowledge transfer is done by computing popularity score, routing score, \emph{HoD} and \emph{HoF}, for each transferred patch (as previously described in Section \ref{scene_transfer}).
In Figure \ref{fig:transf_final}, we also show a qualitative example of the results obtained with or without knowledge transfer.
\smallskip

\begin{table}[!b]
\centering
\caption{(a) Knowledge transfer results on the Stanford-UAV dataset (per-class and overall error). (b) How performance is influenced by the number of trajectories.}
\subfloat[Path forecasting]{
\begin{tabular}[b]{c||cc|c}
\multicolumn{4}{c}{MHD error} \\
\hline
& \emph{Pedestrian} & \emph{Cyclist} & Overall\\
\hline\hline
LP & $34.48$ & $28.09$ & $31.29 \scriptstyle{\pm 1.25}$ \\
PM & $22.75$ & $20.58$ & $21.67 \scriptstyle{\pm 1.19}$\\
IOC~\cite{Kitani2012} & $17.99$ & $18.84$ & $18.42 \scriptstyle{\pm 0.97}$\\
\hline
Ours & $12.36$ & $16.22$ & $\mathbf{14.29} \scriptstyle{\pm 0.84}$\\
\hline
\end{tabular}
} \quad
\subfloat[Impact of training data]{
\includegraphics[width=0.4\textwidth]{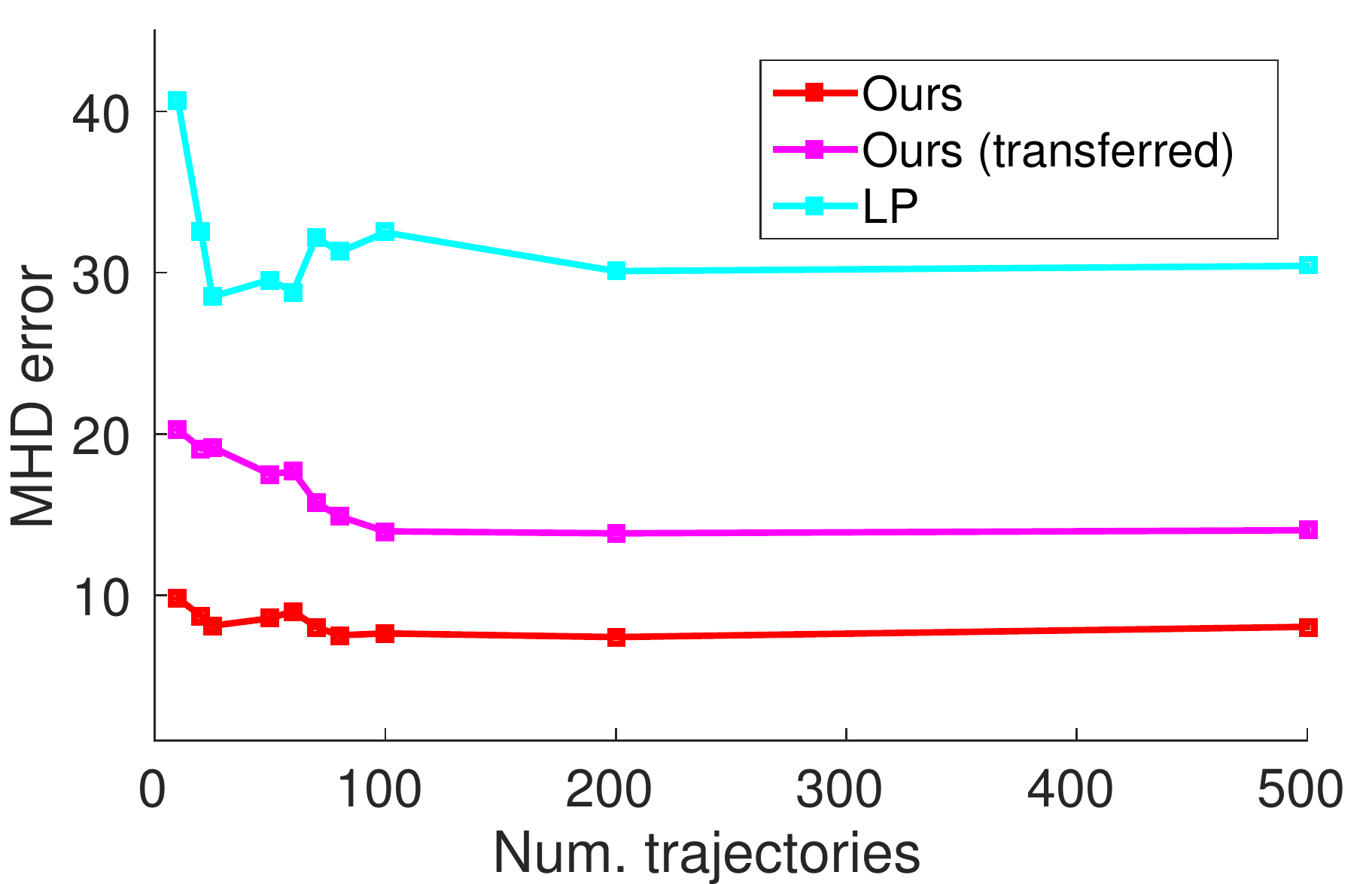}
}
\label{table2}
\end{table}

\textbf{Quantitative results.}
Here we quantitatively evaluate the knowledge transfer capability of our framework.
Therefore we ignore the training trajectories and functional properties encoded in the navigation map of the target scene, and we make predictions using data transferred from $K$ nearest-neighbors retrieved from the training set.
Table \ref{table2} shows that our model after knowledge transfer performs well.
As expected, the predictions obtained starting from the transferred maps are not good as the ones that can be obtained by training from the same scene (i.e. $14.29 \scriptstyle{\pm 0.84}$ \emph{vs} $8.44 \scriptstyle{\pm 0.72}$).
However, we still outperform significantly both the LP baseline and IOC~\cite{Kitani2012}.
We also run an additional baseline based on a simple patch matching scheme (PM), to test the efficacy of our context descriptors. Here each patch is represented with the same visual features used for image parsing, and then these features are used to find the nearest-neighbors.
Finally, it is interesting to note that our performance is significantly better especially for the class \emph{pedestrian}.
We believe this is mainly due to the fact that, in the Stanford-UAV dataset, pedestrians show more non-linear behaviours (while a cyclist on the road has less route choices and so the gain \emph{vs} the simple linear prediction model is less pronounced).
\smallskip

\begin{figure}[!t]
\centering
\subfloat[]{
\includegraphics[width=0.2732\linewidth]{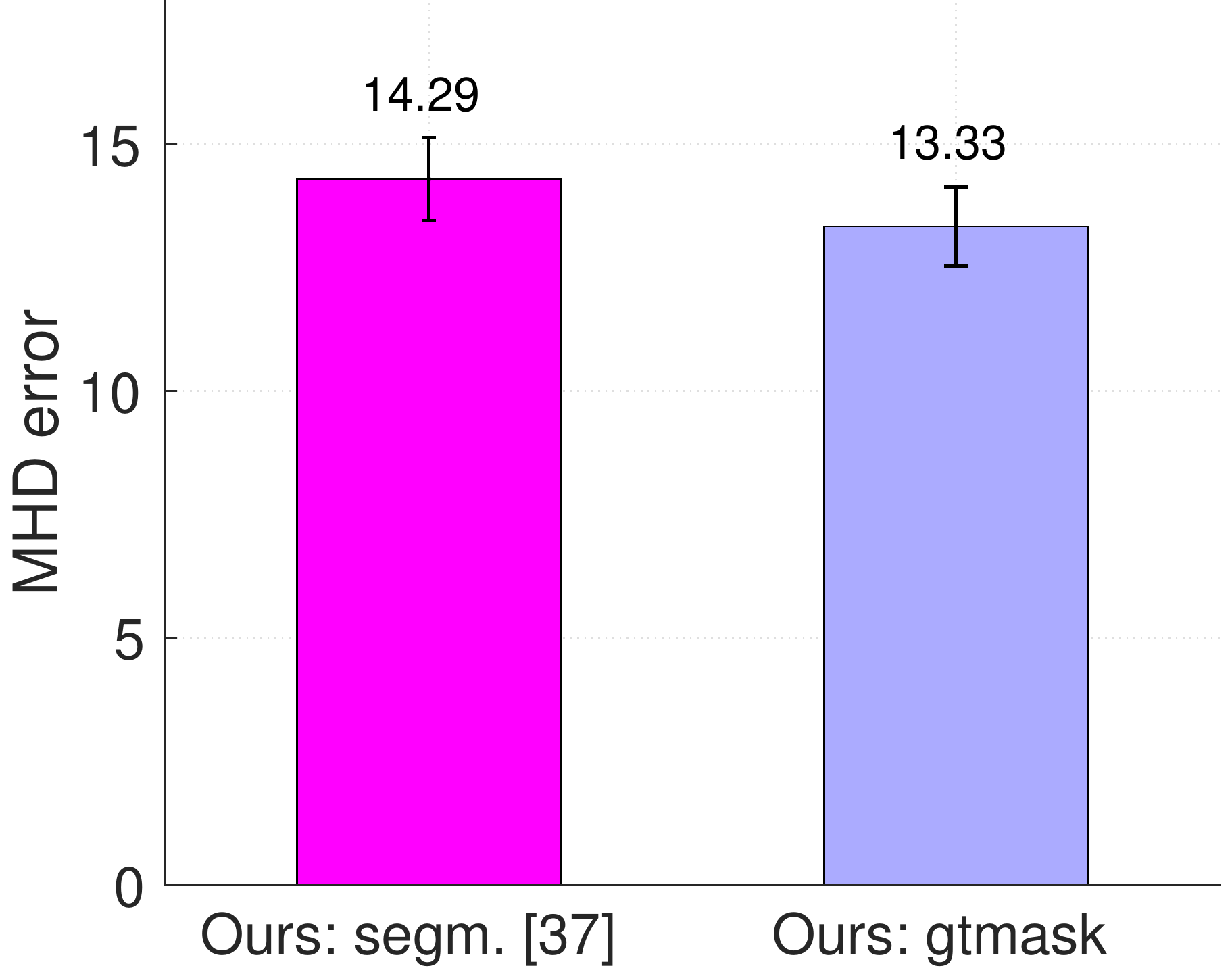}
}
\subfloat[]{
\includegraphics[width=0.2732\linewidth]{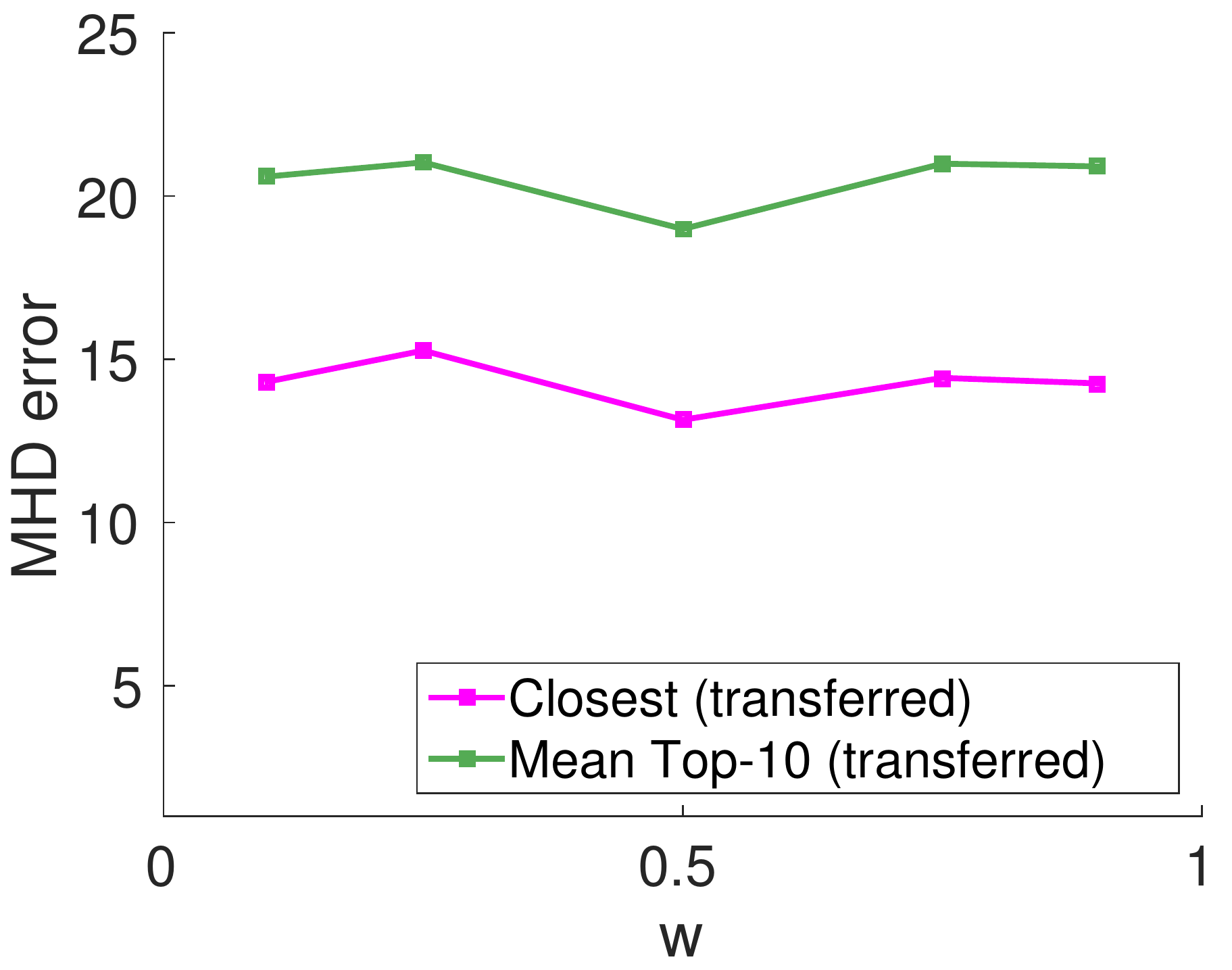}
}
\subfloat[]{
\includegraphics[width=0.2732\linewidth]{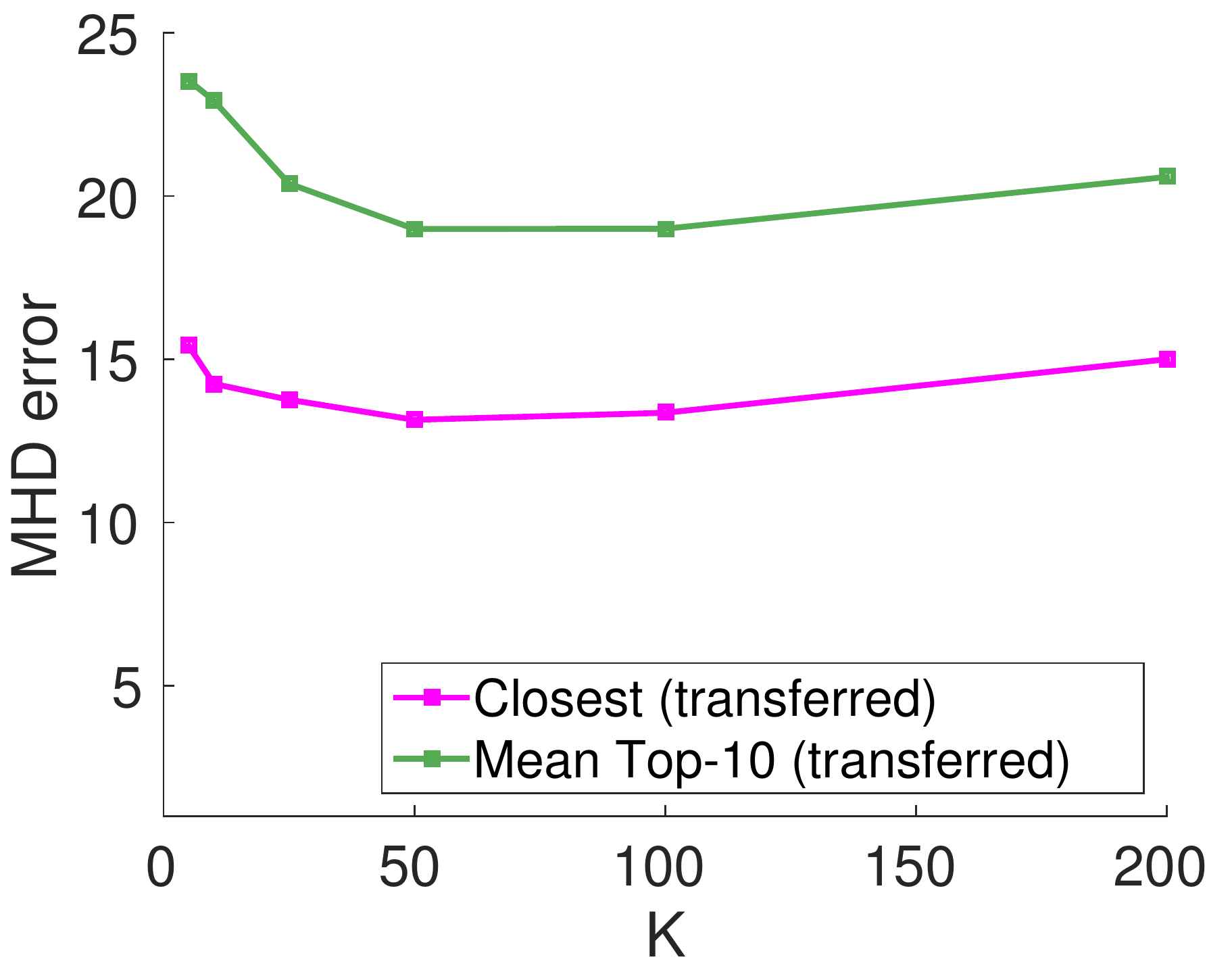}
}
\caption{This figure shows how the performance obtained with knowledge transfer is influenced by the different parameters.}
\label{fig:kt_params}
\end{figure}

\textbf{Impact of the parameters.}
The results reported in Table \ref{table2} has been obtained using $200$ trajectories for training (the plot on the right shows how this number influences performance).
We also evaluate what is the gain that can be achieved using ground-truth segmentation masks, instead of the scene parsing obtained with \cite{Yang2014}.
Interestingly enough, Figure~\ref{fig:kt_params}(a) shows that ground-truth segmentation gives a very slight improvement.
Then the overall robustness of our framework is demonstrated by Figure \ref{fig:kt_params}(b,c). The main parameters of our transfer procedure are $w$ (i.e. the weight of the local and global context features) and $K$, the number of nearest-neighbors used in the retrieval stage. Our best results are obtained with $w=0.5$ and $K=50$.

\section{Conclusions}\label{conclusions}
We introduced a framework for trajectory prediction that is able to model rich navigation patterns for generic agents, by encoding prior knowledge about agent-scene functional interactions of previously observed targets.
Our results show significant improvement over baselines on a large variety of scenes and different classes of target.
More importantly, we show that predictions can be reliably obtained by applying knowledge transfer between scenes sharing similar semantic properties.
Our future work will be focused on modeling simultaneously human-space and human-human interactions to allow predictions in crowded scenes.

\bigskip
\noindent\textbf{Acknowledgements.}
We thank A.~Robicquet for sharing the baseline model of \cite{SUAV,Yama2011}, and H.~O.~Song for helpful comments. This work is partially supported by Toyota (1186781-31-UDARO), ONR (1165419-10-TDAUZ), and MURI (1186514-1-TBCJE). L.~Ballan is supported by an EU Marie Curie Fellowship (No.~623930).


\bibliographystyle{splncs03}
\bibliography{ref}
\end{document}